\documentclass[a4paper]{article}

\usepackage{xcolor}

\usepackage{spconf,amsmath,graphicx}
\ninept

\usepackage{xspace}
\usepackage{url}
\usepackage{pgfplots}

\usepackage{amssymb}
\usepackage{pifont}

\definecolor{tableau_blue}{RGB}{31, 119, 180}
\definecolor{tableau_orange}{RGB}{255, 127, 14}
\definecolor{tableau_green}{RGB}{44, 160, 44}
\definecolor{tableau_red}{RGB}{214, 39, 40}
\definecolor{tableau_purple}{RGB}{148, 103, 189}
\definecolor{tableau_brown}{RGB}{140, 86, 75}
\definecolor{tableau_pink}{RGB}{227, 119, 194}
\definecolor{tableau_gray}{RGB}{127, 127, 127}
\definecolor{tableau_olive}{RGB}{188, 189, 34}
\definecolor{tableau_cyan}{RGB}{23, 190, 207}

\newcommand{\datasetname}{The Edinburgh International Accents of English Corpus}

\newcommand{\datasetnameshort}{\texttt{EdAcc}\xspace} 

\begin{document}

\title{The Edinburgh International Accents of English Corpus: \\Towards the Democratization of English ASR}
\name{Ramon Sanabria, Nikolay Bogoychev, Nina Markl, Andrea Carmantini, Ondrej Klejch, Peter Bell}

\address{
  School of Informatics, The University of Edinburgh}

\maketitle
\begin{abstract}
English is the most widely spoken language in the world, used daily by millions of people as a first or second language in many different contexts. As a result, there are many varieties of English. 
Although the great many advances in English automatic speech recognition (ASR) over the past decades, results are usually reported based on test datasets which fail to represent the diversity of English as spoken today around the globe.
We present the first release of \datasetname\ (\datasetnameshort). This dataset attempts to better represent the wide diversity of English, encompassing almost 40 hours of  dyadic video call conversations between friends. Unlike other datasets, \datasetnameshort\ includes a wide range of first and second-language varieties of English and a linguistic background profile of each speaker. 
Results on latest public, and commercial models show that \datasetnameshort\ highlights shortcomings of current English ASR models. The best performing model, trained on 680 thousand hours of transcribed data, obtains an average of 19.7\% word error rate (WER) -- in contrast to the 2.7\% WER obtained when evaluated on US English clean read speech. Across all models, we observe a drop in performance on Indian, Jamaican, and  Nigerian English speakers. Recordings, linguistic backgrounds, data statement, and evaluation scripts are released on our website under CC-BY-SA\footnote{\url{https://creativecommons.org/licenses/}} license.\footnote{\mbox{\url{https://groups.inf.ed.ac.uk/edacc/}}}
We hope that this work will encourage future research on a wider range of English varieties to create more accessible speech technologies.
\end{abstract}

\section{Introduction}

English is a first language for more than 370 million people \cite{campbell2008ethnologue}, having been spread through (settler) colonialism over hundreds of years \cite{Bauer2003IntroductionInternationalVarieties}. In recent decades, English has only gained power as a lingua franca in global business, international politics, media and pop culture, and academia. As a result, there are an estimated 1 billion people who speak English as a second language and most of the state-of-the-art language technology research caters to it.

Even though language technologies work better for English than for other languages, there are still vast performance differences between English varieties, with higher performance for US and UK \cite{KoeneckeEtAl2020Racialdisparitiesautomated,BlodgettEtAl2020Languagetechnologyis,Markl2022LangVar}. There are hundreds of varieties of English spoken by people in different geographical areas and social contexts \cite{VanHerk2018Whatissociolinguistics}. Most of these are poorly supported by automatic speech recognition systems (ASR). Furthermore, even though there is research on supporting ``accented English''\footnote{A slightly confusing, if established, term since every speaker of English (or any language) has an accent.} most ASR development is focused on US, and UK language speakers of English.

More concretely, in recent years, there have been notable advances in English language ASR. These advances are usually reported as word error rates (WER) on established benchmark datasets. The current state-of-the-art on Switchboard \cite{godfrey1992switchboard} -- an US English conversational  speech dataset  -- is 4.3\% WER \cite{tuske2021limit}. With further advanceces in self-supervised learning \cite{baevski2020wav2vec}, a WER of 1.4\%, and 3.2\% was obtained, respectively on Librispeech  \cite{panayotov2015librispeech} -- an US English read speech dataset. However, the robustness of these results has been questioned by experiments showing  that they are not actually representative in other English variations \cite{hsu2021robust,lin2022analyzing}.  Taken together, these findings suggest a lack of corpora that accurately represents the depth and breadth of English varieties in conversational settings.

Most public datasets do not represent the diversity of English and English speakers in the real world. We observe that they cover outdated or over-explored domains (eg., narrowband telephone speech, or read speech), and varieties (eg., US English). Language change, just like language variation, is an inherent and natural feature of language, and therefore older datasets further risk becoming unrepresentative of current language use -- a problem relevant to all fields of natural language processing, eg., \cite{bender2021dangers}. Additionally, current datasets lack detailed documentation about the speakers and their linguistic backgrounds \cite{markl-2022-mind}, making it hard to draw conclusions on the reported results.

We introduce the first release of an ongoing project \datasetname\ (\datasetnameshort). Our dataset contains almost 40 hours of video call dyadic English language conversation between speakers who know each other. The conversations range in duration from 20 to 60 minutes. \datasetnameshort is diverse, containing more than 40 self-reported English accents from speakers from 51 different first languages. We also collected their linguistic background including any languages they speak, how long they have spoken English, and places where they have lived for extended periods of time. We release our dataset with the responses from each speaker.

The self-reported statistics and qualitative and quantitative analysis show that \datasetnameshort is linguistically diverse and challenging to current English ASR systems. Our extensive experimentation on open source state-of-the-art models pre-trained and fine-tuned on a variety of publicly available datasets shows that latest research on ASR does not generalize to many L1 and L2 English speakers. Despite being trained on quantity of data that would be prohibitively large for most researchers, the best performing model \cite{radford2022robust} achieves 19.7\% WER, far from the reported performances on standard datasets.

Overall, our results show that more research is required in order for English ASR systems to generalise to other variants of English. We hope \datasetnameshort encourages future research on speech processing for a wider range of English accents.

\section{Related Work}

The main source of motivation for this work is the lack of ASR datasets which go beyond L1 varieties of English. TIMIT \cite{garofolo1993darpa}, and the Wall Street Journal \cite{paul1992design} are some of the earliest datasets for English language ASR. Both of them are exclusively composed of American English read speech. The DARPA-sponsored Switchboard dataset \cite{godfrey1992switchboard} contains telephone conversations by speakers from several dialect regions of American English. Although the domain of phone conversations leads to more expressive language, it is not representative of modern remote conversational medium of the video call. Moreover, phone conversations are narrow band which limits frequency resolution of the speech signal. Later on, Librispeech \cite{panayotov2015librispeech} and its extension \cite{kahn2020libri} increased the amount of training data, but the domain -- American English read speech -- remain the same.   In contrast, \datasetnameshort\ is composed by more natural interactions (conversation between friends), higher quality (wide band) and updated domain (video call conversation).

Recent speech resources focused on wider variants of English \cite{demirsahin2020open,zhao2018l2}. However, only The Accented English Speech Recognition Challenge
(AESRC2020), and Mozilla Common Voice (MCV) \cite{ardila2019common,shi2021accented} are comparable to our work. Both datasets have a similar data collection framework: speakers read, and record sentences using a computer or a mobile devices. MCV is focused on collecting multiple languages, and AESRC2020 is specifically targeted to English accents. Both of them have a higher audio quality than previous datasets, and collect different English accents. However, neither of them annotates the speaker's accent (or linguistic background) properly. MCV let speakers report their own accent, and AESRC2020 uses speaker country of origin -- both methods lacks precision in terms of accent definition. Perhaps most significantly, in contrast to \datasetnameshort, both datasets are composed by read speech which limits the naturalness of speech, and can mask the accent. 

Our dataset also relates to recent work on predictive bias in ASR. In the US, commercial systems have been shown to perform much worse for speakers of African American English than white speakers of US English \cite{KoeneckeEtAl2020Racialdisparitiesautomated,MartinTang2020Understandingracialdisparities}. Disparities between performance for different varieties of English have also been reported in British English ASR \cite{Markl2022LangVar}, and for global varieties \cite{MeyerEtAl2020Artiebiascorpus}. We designed \datasetnameshort as as a tool to identify such biases, and facilitate the research towards a solution for this problem.

\section{The Edinburgh International Accents of English Corpus}

\subsection{Data Collection}
Our data collection process is designed to provide a simple framework for eliciting naturalistic speech by allowing speakers to record relaxed conversations using the Zoom video call software. A questionnaire distributed to participants further enables the curation of a well-documented and diverse dataset. 


Participants were initially recruited through the authors' personal and professional (local and global) networks. As the data collection progressed, speakers were also recruited through an online micro-work platform.\footnote{ \mbox{\url{https://www.fiverr.com/}}} Each Participant was compensated with 10 GBP for every 15 minutes of conversation. We selected participants according to their linguistic background to encourage as much diversity as possible.

To capture participants' linguistic backgrounds we asked them about: any first languages (acquired before age 5), any second languages, when they started learning English, which language they mostly use in different domains (work, friends, family) and any places where they have lived for more than three years. We also asked them how long they have known their conversation partner and whether they usually speak English with them. Finally, we asked them to self-describe their accent in English. To capture their social background we also asked about their age, gender, ethnic background and education. Find the specific questions in the dataset statement.$^2$

 The use of video call software made our approach scale-able and simple: conversations could be recorded at the same time in multiple places, allowing us to reach speakers in different parts of the world. As a side issue, due to software limitations, only one audio audio channel could be recorded -- instead of one channel for each speaker. Conveniently, this setting also replicates real-world acoustic condition where ASR engines are usually deployed. The contributors were provided with detailed instructions on how to record the conversation (on audio, and if they wished to do so, video\footnote{We are not publishing video data for this version of the dataset. We will do it in next releases.}). We provided some discussion prompts about topics such as hobbies. This design is inspired by data collection procedures in sociolinguistics \cite{schilling_2013,VanHerk2018Whatissociolinguistics}, where an engaging topic can reduce self-consciousness and promote more natural speech patterns. Informal speech is further promoted by the interlocutor -- all participants talked with friends or acquaintances. To a certain extent, this design may also limit linguistic accommodation effects, though it should be noted that all interactions involve some accommodation or alignment \cite{giles_coupland_coupland_1991}. Finally, the ``observer's paradox'', where participants in an experiment feel self-conscious and adjust their speech, is further reduced by asking participants to self-record their conversations \cite{schilling_2013,VanHerk2018Whatissociolinguistics}.  Before starting the conversation, each speaker was also asked to read the same control passage\footnote{The ``Stella'' passage designed to elicit a wide range of features of different English accents. It was developed and used by the Speech Accent Archive \cite{Weinberger2015}.} to allow evaluation on a controlled domain and enable detailed linguistic analysis.

\datasetnameshort\ is designed specifically to make our data compliant with CC-BY-SA so that data can be fully shareable. Each speaker was given an speaker ID, and asked to identify themselves with it during the conversation. We manually verified that no sensitive data about the speakers or anyone else was shared during the conversation. In the final step of the data collection, participants signed a consent form with the data protection statement and confidentiality policy, and then received their compensation for their contribution. We believe that the transparent design of our collection and distribution pipeline makes it secure for data subjects.

All conversations were manually transcribed by multiple professional transcribers. Each turn was manually segmented and orthographically transcribed, including overlaps between speakers, noise, laughter and hesitations. The transcription company removed the stored data ten days after receiving it. 

Once the transcriptions were ready, we post-processed them so they can be used for evaluation. We made them consistent with orthography, and lexicon used in public systems -- we removed punctuation, and normalized ambiguous spelling. We localized sounds (eg., laughs) and words that a system can transcribe in multiple manners (eg., numbers) and ignore them during scoring. We localize them by force-aligning the speech and the transcripts using Kaldi \cite{Povey_ASRU2011}. We split the data on development and test sets randomly using  Kaldi's \texttt{subset\_data\_dir\_tr\_cv.sh} script resulting in 31, and 30 conversations each. There is no speaker nor conversation overlap between sets.

\subsection{Accent Descriptions}

A trained linguist slightly standardised the contributors' accent description to enable us to compare performance across groups of speakers (Section~\ref{sec:exp})\footnote{Note that the public release of the dataset does not contain any normalization.}. To do this, some specific accent descriptors were simplified (eg., ``Scottish (Fife)'' to ``Scottish English''), some broader descriptors were mapped to a more commonly used descriptor (eg., ``American accent'' to ``US English''). Some generic descriptors (eg., ``fluent'') were mapped to a local descriptor based on information about the participant's location history and linguistic background. We want to stress that these labels are not definitive. Differences between ``accents'' are contested and an issue of identity as much as one of phonetics for first and second language speakers. The dataset is roughly balanced between male and female speakers but other gender groups are underrepresented. Speakers range in age from 18 to 65 years (mean age 30) and contributors also provided information about their ethnic background, linguistic background, location history and education. You can find more detailed statistics in our Data Statement on our web page.$^2$

\section{Experiments}
\label{sec:exp}

\subsection{Automatic Speech Recognition Models}

To evaluate ASR performance on \datasetnameshort, we test three different model classes: Wav2vec2.0 \cite{baevski2020wav2vec}, Whisper \cite{radford2022robust}, and  a commercial engine from an (anonymized) well-established company. 

\textbf{Whisper} is  a traditional encoder-decoder-based system trained on unreleased 680,000 hours of multilingual and multitask data collected from the web. In contrast to new self-supervised architecture, Whisper is trained exclusively in a weakly supervised manner. To process long utterances, the system segments the audio in 30 second chunks.

\textbf{Wav2vec2.0} is a pre-trained self-supervised-based encoder, that we fine-tuned with CTC loss on transcribed datasets. During the first stage, the encoder is pre-trained to predict the quantized representation of masked segments of speech. After that, the encoder is fine-tuned on sequences of characters. During decoding, we integrate a 4-gram language model (LM) \cite{kenlm} that constrains the search over likely sequences of words. To make our evaluation consistent with Whisper we decode 30 seconds of audio at a time\footnote{We also experimented using rVAD \cite{tan2020rvad}, an unsupervised voice activity detection system, for segmentation and under-performed 30 second segmentation on the development set.}. We experiment with two encoders pre-trained on different datasets; Wav2vec2.0 (pre-trained on Libripeech), and Robust Wav2vec2.0 \cite{hsu2021robust} (pre-trained on MCV, Libri-light, and Switchbord). We test each encoder fine-tuned on Librispeech, Switchboard, AMI \cite{mccowan2005ami}, and MGB \cite{bell2015mgb}, and combine them with language models trained in these same datasets.

\textbf{Commercial system} belong to a known (anonymized) company. This engine is as a black-box, with model architecture and training data undisclosed. The company offers different models tuned to recognize speech from particular accents -- their system automatically selects the best suited model for each conversation. We experimented by manually selecting models for each conversation, but it did not have any notable effect on the results.

\begin{table}
\vspace*{-\baselineskip}
\begin{center}
\begin{tabular}{ c c c c c c}
Model               & \datasetnameshort\ dev     & \datasetnameshort\ test &  test-clean     & test-other   \\    \hline 
W2V2.0          & 33.4     &    36.1   & 2.9     & 5.6      \\  
Company             & 17.9    & 18.7       & 3.8     & 7.4         \\  
Whisper             & 16.4     & 19.7      & 2.7     & 5.6         \\  
\end{tabular}
\end{center}
\caption{Results of selected systems on the \datasetnameshort's test, and development set, and Librispeech test sets on the three selected ASR models.}\label{tab:general_result}
\end{table}

\subsection{Quantitative Analysis}

\begin{figure}[]
    \centering
    \begin{tikzpicture}
        \begin{axis}[
            ybar,
            bar width=3pt,
            axis y line*=left,
            axis x line*=bottom,
            y label style={at={(axis description cs:0.1,0.5)},anchor=south},
            ylabel={WER},
            symbolic x coords={
                S. African English,
                Ghanain English,
                Irish English,
                Scottish English,
                US English,
                Southern British,
                Indian English,
                Jamaican English,
                Nigerian English
            },
            xtick=data,
            height=6cm, 
            width=0.5\textwidth, 
            enlarge x limits={0.1},
            xticklabel style={
            column sep=10pt,  
            rotate=45, 
            anchor=north east 
            },
            legend style={/tikz/every even column/.append style={column sep=0.25cm}, draw=none, at={(0.55,1.05)},anchor=north,
            legend columns=-1, legend cell align={left}},
            enlarge x limits={abs=0.75cm},
        ]
                \addplot[fill=tableau_red] coordinates {
            (S. African English, 12.49)
            (Ghanain English, 12.9)
            (Irish English, 14)
            (Scottish English, 14.84)
            (US English, 16.084)
            (Southern British, 16.59625)
            (Indian English, 19.74333333)
            (Jamaican English, 21.45666667)
            (Nigerian English, 22.92)
        };
        \addplot[fill=tableau_blue] coordinates {
            (S. African English, 13.43)
            (Ghanain English, 12.72)
            (Irish English, 16.53666667)
            (Scottish English, 17.135)
            (US English, 17.318)
            (Southern British, 18.91375)
            (Indian English, 20.40333333)
            (Jamaican English, 26.00666667)
            (Nigerian English, 25.635)
        };
        \addplot[fill=tableau_orange] coordinates {
            (S. African English, 24.52)
            (Ghanain English, 23.44)
            (Irish English, 30.21)
            (Scottish English, 31.495)
            (US English, 32.016)
            (Southern British, 37.83375)
            (Indian English, 39.70666667)
            (Jamaican English, 46.19)
            (Nigerian English, 54.04)
        };
        \legend{Whisper, Commercial Model , Wav2vec2.0}
        \end{axis}
    \end{tikzpicture}
\caption{WER of selected systems on conversations from the development set of \datasetnameshort where both speakers has the same English variety.}
    \label{fig:en_results}
    \label{fig:grouped_bar_chart}
\end{figure}
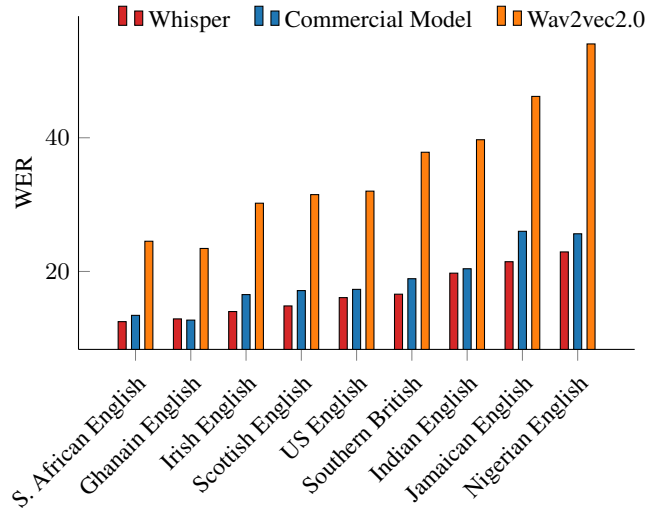

We start by measuring the general complexity of \datasetnameshort\ by computing WER on development, and test sets at the conversation level.\footnote{Current versions of the dataset have sentence level alignments} We only report results on one model for each group. We select them based on their performance on the development set. For Wav2vec2.0 we use an encoder pre-trained on Libri-light, MCV, Switchboard, and Fisher, fine-tuned on Librispeech with a LM trained on MGB. For Whisper, we use the \texttt{large} model without conditioning on the previously decoded text. Table \ref{tab:general_result} shows the \datasetnameshort's test and development set results on the selected models. We observe that the commercial model, and Whisper outperform Wav2vec2.0 by a large margin, which might be due being exposed to more, and more diverse English data. More importantly though, these results suggest a poor fit of academic scale models to realistic English data. 


Next, we want to see whether \datasetnameshort reveals blindspots that Librispeech does not capture. We do this by comparing WER between both datasets on all three models. Table \ref{tab:general_result} shows this comparison on the \texttt{test-clean}, and \texttt{test-other} sets. We observe a considerable drop in performance in Wav2vec2.0 when comparing Librispeech, and \datasetnameshort results. This gap indicates a worrying lack of robustness of the model when exposed to a real world setting. Although this gap is still present in other models, the difference is not as stark, which indicates higher robustness. We hope these results encourages future accent robustness research on academic-size models.

Until now, we have discussed global WER, and compare it across datasets. Now, we will make use of the linguistic background reported by the speakers to discern the performance on different accents. Although we could show performance on many speaker dimensions, we decided to only report WER on first language varieties represented in the dataset: South African English, Ghanaian English, Irish English, Scottish English, US English, Southern British English, Indian English, Jamaican English and Nigerian English. The L2 speakers of English in the dataset vary in terms of their first languages and how long, how and where they have learned English. As a result it is more difficult to compare within and across L2 speaker groups. We leave this for future work. Because we can only compute one WER for each conversation, we limit this analysis to conversations where both speakers use the same English variety. Figure \ref{fig:en_results} shows this comparison. We see a considerable performance gap between Wav2vec2.0, and other models across all L1 varieties. Consistent with previous work \cite{MeyerEtAl2020Artiebiascorpus,Markl2022LangVar}, we observe considerable drop in performance on specific L1 varieties, such Jamaican, Indonesian, Nigerian, and Kenyan English. These differences are particularly noticeable in Wav2vec2.0 models, which stress the urgency to tackle these problems. We are happy to see that an open-source model (Whisper), and a commercial model have a more consistent performance across accents. 

\subsection{Qualitative Analysis}

\begin{figure}\ 
\vspace*{-\baselineskip}
    \centering
    \includegraphics[width=0.9\columnwidth]{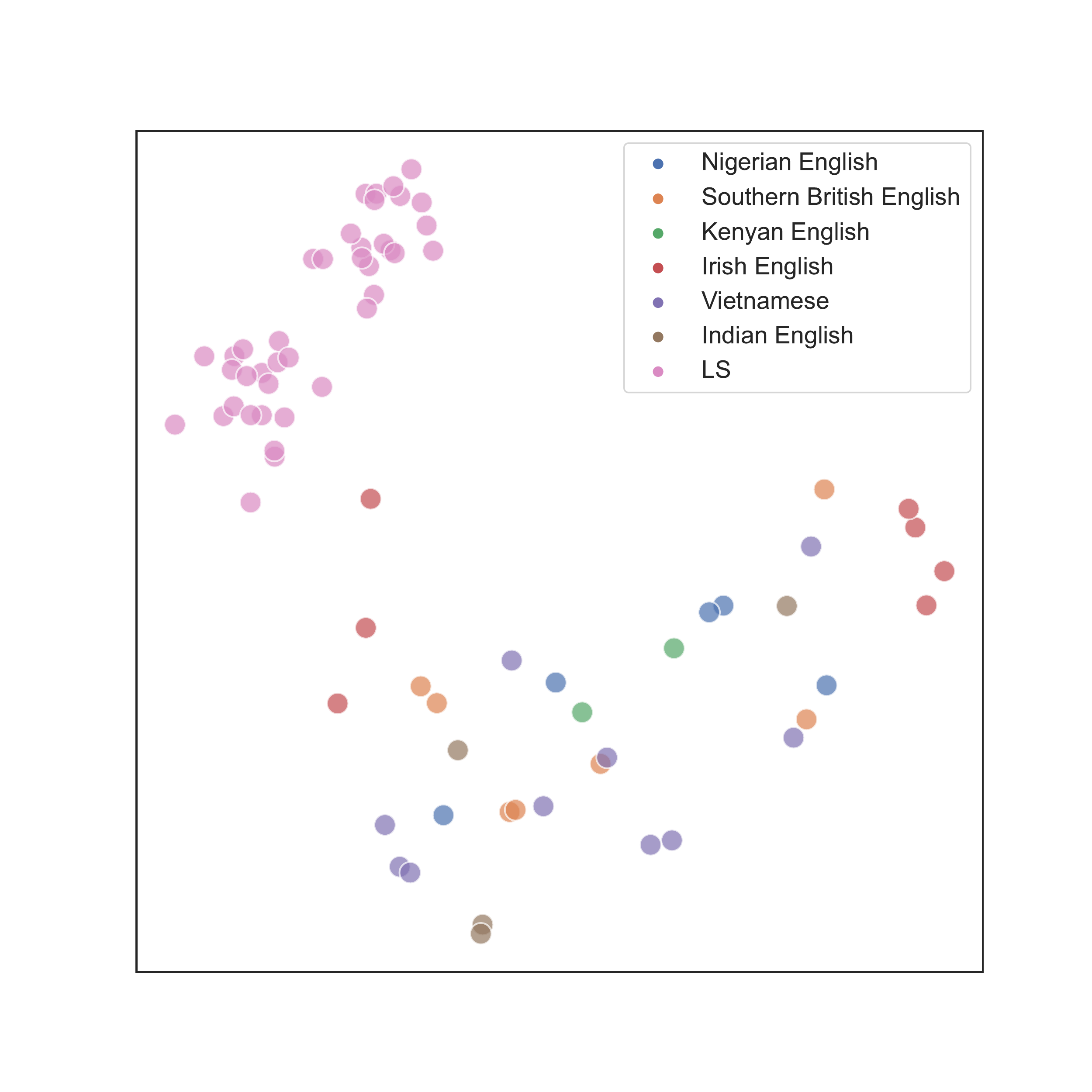}
     \caption{t-SNE speaker visualization. We averaged the utterance-level language classification embedding \cite{desplanques2020ecapa} from 10 random utterances from speakers with different English variations. In pink, we sample several utterance from the test-clean Librispeech.}
    \label{fig:embeds}
     \vspace*{-\baselineskip}
\end{figure}

Our self-reported forms, and quantitative results indicate that \datasetnameshort\ has a wide English diversity -- diverse reported accents, different WER between accent groups, and large performance gap compared to traditional English datasets. As accents cannot be straightforwardly mapped to a ``true'' label\footnote{As noted above, accents and how people describe them, are deeply linked to social identities. Our dataset is furthermore focused on people who may or may not speak English as a first language and many of them have lived in different places throughout their life. As a result mapping any one speaker to any one clearly defined accent is impossible.}, we propose to ``objectively'' analyze the diversity of \datasetnameshort\ by visualizing embeddings extracted from a language identification model. We may be able to visualise accent variation using a language ID embeddings because the English accent of many speakers is usually influenced by the phonological system of their first language.  For this experiment we use SpeechBrain's \cite{speechbrain} ECAPA-TDNN model \cite{desplanques2020ecapa} trained to recognize 107 languages from spoken utterances. Figure \ref{fig:embeds} shows a 2 dimensional t-SNE representation of the 6 most common accents in the dataset. For each speaker we randomly sample and average the representation of 100 utterances. For a consistent comparison, we also sample multiple speakers from Librispeech. We observe that \datasetnameshort\ has a wider diversity in the embedding space than Librispeech. This analysis suggest that our dataset is linguistically more diverse and explores a space not covered by Librispeech -- the standard set for the speech recognition the community.

Apart from linguistic diversity, a dataset must show some structure, so that future research can experiment with specific groups -- accents in our case. Figure \ref{fig:embeds} shows an underlying structure -- speakers with the same accent are clustered together, and some phonetically similar accents are closer than others. Concretely, we observe that Vietnamese and Indian form two clusters -- except for some outlyers. We can see that Nigerian, and Kenyan English -- two similar variants -- are close together in the embedding space. Overall, these observations indicate that \datasetnameshort is not only diverse but also linguistically structured.

\section{Conclusions and Future Work}

We present the first release of \datasetname\ (\datasetnameshort) a new automatic speech recognition (ASR) dataset composed of 40 hours of English dyadic conversations between speakers with a diverse set of accents. To facilitate error analysis on specific English accents, we also provide a detailed linguistic profile for each speaker containing their first language, years speaking in English, years lived in an English-speaking country and more. Results on different state-of-the-art systems show that \datasetnameshort\ is generally challenging, and can identify problems on specific English accents (eg., Jamaican, and Kenyan English). Qualitative results show that \datasetnameshort is more diverse than traditional datasets, and covers more linguistic variation.

We found that assigning an accent to each speaker is difficult. At the same time, having speakers grouped in these categories is important to spot specific problems in ASR systems and construct a balanced dataset. Therefore, a promising direction for future work includes exploration of accent classification or clustering strategies to make future iterations of  \datasetname\ diverse and balanced.

\section{Ethical Considerations}\label{sec:ethics}

This project has been approved by the University of Edinburgh, Informatics Ethics Board -- Ref. 49776. It has been funded by the Institute for Language, Cognition, and Computation at the University of Edinburgh. All contributors provided informed consent to publicly share their speech recordings and demographic information and were paid 10 GBP for every 15 minutes of conversation. We acknowledge that despite efforts to design an accessible and fair data curation process, we are only representing a small section of all English speakers. In future iterations of the project, we will keep encouraging speakers from underrepresented groups to contribute.

\vfill
\pagebreak

\ninept
\bibliographystyle{IEEEtran}
\bibliography{mybib}

\end{document}